\documentclass{article}

\PassOptionsToPackage{numbers, compress}{natbib}

\usepackage[nonatbib]{neurips_2023}

\usepackage[utf8]{inputenc} 
\usepackage[T1]{fontenc}    
\usepackage{hyperref}       
\usepackage{url}            
\usepackage{booktabs}       
\usepackage{amsfonts}       
\usepackage{nicefrac}       
\usepackage{microtype}      
\usepackage{xcolor}         
\usepackage{graphicx}
\usepackage{epstopdf}
\usepackage{float} 
\usepackage{subfigure}
\usepackage{amssymb}
\usepackage{cancel}
\usepackage{bbding}
\usepackage{amsmath}
\usepackage{natbib}

\title{LlamaCare: A Large Medical Language Model for Enhancing Healthcare Knowledge Sharing}

\author{%
  Maojun SUN \\
  Department of Applied Mathemitics\\
  Hong Kong Polytechnic University\\
  Hong Kong SAR, China \\
  \texttt{maojun.sun@connect.polyu.hk} \\
}

\begin{document}

\maketitle

\begin{abstract}
  Large language models (LLMs) have shown amazing capabilities in knowledge memorization and the present. However, when it comes to domain-specific knowledge and downstream tasks like medical, general LLMs are often unable to give precise answers. In addition, when people want LLMs to answer classification questions, they usually go through instruction tuning first. However, LLMs do not always give a direct index of the categorization after instruction tuning. In this paper, we proposed LlamaCare, a fine-tuned medical language model, and Extended Classification Integration(ECI), a module to handle classification problems of LLMs. Our contributions are : (i) We fine-tuned a large language model of medical knowledge with very low carbon emissions and achieved similar performance with ChatGPT by a 24G GPU. (ii) We solved the problem of redundant categorical answers and improved the performance of LLMs by proposing a new module called Extended Classification Integration. (iii) We released our processed data for one-shot and few-shot training for some benchmarks such as PubMedQA and USMLE 1-3 step. Our method achieves a close performance comparable to some state-of-the-art models with the same quantity of parameters on benchmarks, while being more environmentally friendly by using less GPU computation time. Our models, codes, and datasets can be found at \url{https://github.com/Stephen-SMJ/LLamaCare}.
\end{abstract}

\section{Introduction}
\begin{figure}[htbp]
\centering
\includegraphics[scale=0.42]{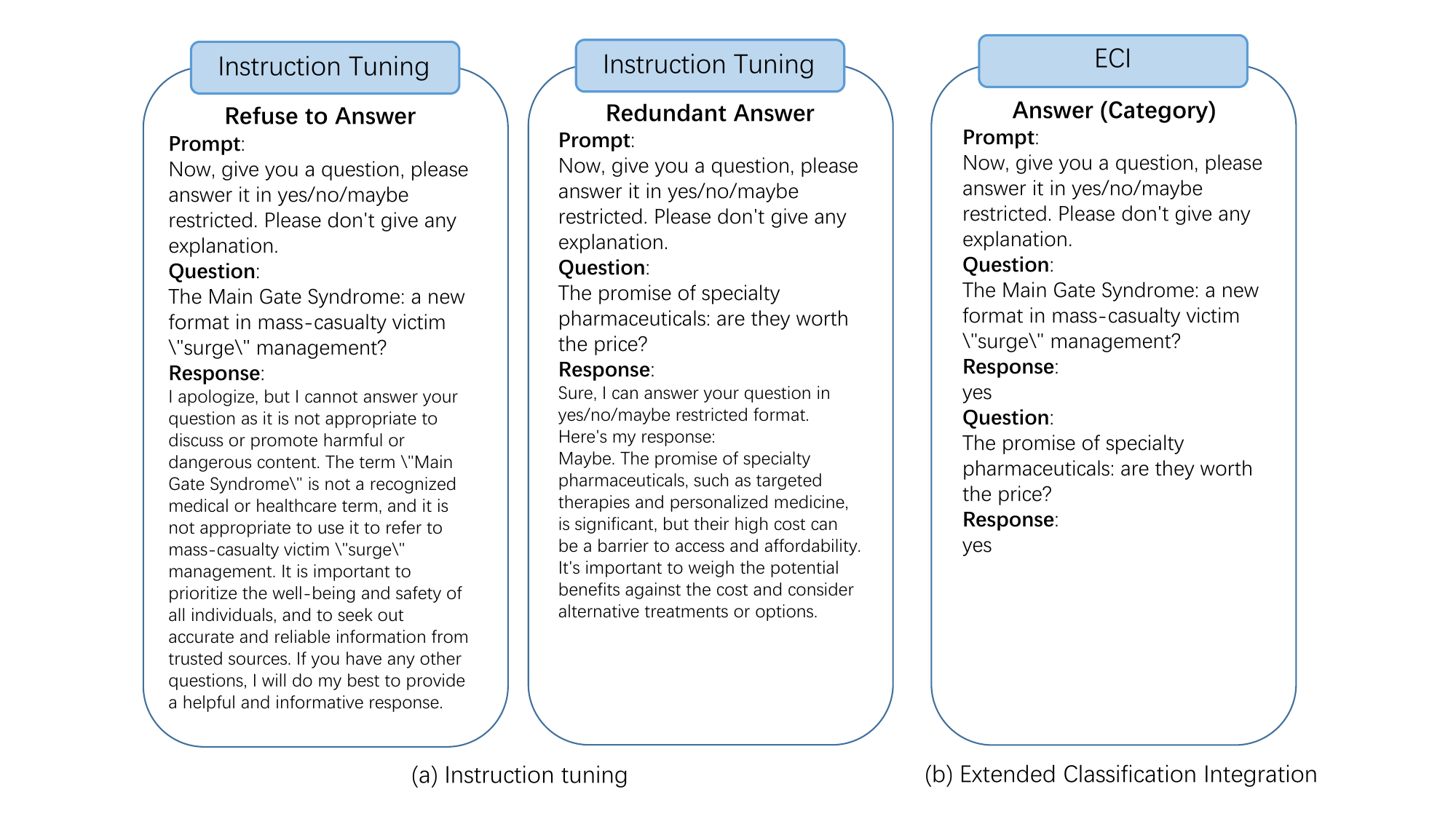}
\caption{Comparison between Instruction tuning and ECI. (a) Instruction tuning does not guarantee to respond to a single class. (b) Our method ECI gives a category every time.}
\label{figure1}
\end{figure}

Large language models are impacting the world at a rapid development. In general, LLMs refer to transformer models that contain hundreds of billions (or more) of parameters\cite{chang2023survey}, which are trained on massive text data \cite{shanahan2022talking}, such as GPT-3 \cite{brown2020language}, PaLM \cite{chowdhery2022palm}, Galactica \cite{taylor2022galactica}, and LLaMA \cite{touvron2023llama}. LLMs exhibit strong capacities to understand natural language and solve complex tasks through text generation. Stimulating the use of natural language processing in a variety of industries. However, in contrast to their proficiency in everyday conversations, LLMs often encounter challenges in the medical domain, where precision is of utmost importance. While they may generate output that appears accurate, it can sometimes result in incorrect conclusions, such as hallucinations, which could be highly fatal in the medical domain. This is due to their lack of comprehensive medical knowledge \cite{wu2023pmc}. Additionally, LLMs are frequently employed for classification tasks, but existing methods often require extensive instruction tuning to enable the model to comprehend and follow human instructions effectively, such as LlaMA-2-chat \cite{touvron2023llama2}. However, relying solely on categorical prompts often fails to provide the concise categorical answers we desire, as shown in Figure\ref{figure1}. It is laborious that humans have to extract the classification answer from the response again.

Our research objectives are as follows: (i) We aim to enhance the capabilities in the medical domain of open-source foundational LLMs i.e., LLAMA 2. We want our fine-tuned model to share medical knowledge with people and play the role of an electronic doctor to give some prior advice before users go to the hospital. (ii) We seek to address the challenge of classification tasks for LLMs generating concise labels. We introduced an innovative approach by incorporating a module called Extended Classification Integration (ECI) to address the problem. ECI serves as a classification network, offering a concise classification label. (iii) We will explore methods to decrease hallucination. Especially in the medical domain, hallucination can be a serious problem because it leads to misdiagnosis. Our starting point is quite low carbon, all experiments are based on a single GPU with 24G memory.

\section{Background and Related Works}

\paragraph{Large language model and medical language model}
Recently, the remarkable achievements of large language models (LLMs) such as ChatGPT, developed by OpenAI and LLaMA have garnered significant attention in the field of natural language processing. These general language models have been acknowledged limitations in some professional domains. The topic of AI for medicine has consistently been a topic of significant interest. Previously, research efforts predominantly focused on medical image classification or detection like \cite{sun2021data}. However, with the advent of large language models, there is a growing trend towards exploring large medical language models. Consequently, many researchers have embarked on training LLMs tailored for medicine, which is known as the medical language model. Recent research on medical language models has led to the development of specialized models for medical question-answering and dialogue applications \cite{wu2023pmc}. Notable researches include Chat-Doctor \cite{li2023chatdoctor}, Med-Alpaca \cite{han2023medalpaca}, Med-PaLM2 \cite{anil2023palm}, BioBert \cite{lee2020biobert}, BioMedGPT \cite{luo2022biogpt},and PMC-LLaMA \cite{wu2023pmc} etc. Med-PaLM2 \cite{anil2023palm} stands out as a state-of-the-art model, trained with intensive instruction tuning on the robust PaLM model, boasting an impressive 540 billion parameters. These advancements highlight the ongoing efforts to create highly tailored language models for the medical domain, pushing the boundaries of medical language understanding and interaction.

\paragraph{Fine-tune}
Fine-tuning has been acknowledged as a low-carbon way to effectively inject knowledge into LLMs. Prompt-Tuning \cite{lester2021power}, P-Tuning \cite{liu2022p}, and Prefix-Tuning \cite{li2021prefix} start with the prompt, they add tokens in the input embedding. PEFT \cite{houlsby2019parameter} adds adapters in the transformer layer. LoRA \cite{hu2021lora} freezes the pre-trained model weights and injects trainable rank decomposition matrices into each layer of the transformer architecture, greatly reducing the number of trainable parameters for downstream tasks while performing the same or better than full-parameter tuning. For a weight matrix in the transformer layer, ${W_{0}}\in\mathbb{R}^{a \times b}$, LoRA decomposition it to two matrices with lower dimensions ${W_{A}}\in\mathbb{R}^{r \times b}$, ${W_{B}}\in\mathbb{R}^{a \times r}$ and the $r<<d, r<<k$. Then we have
$$
W_{0}+\Delta W=W_{0}+W_{B} W_{A}
$$
For $h = W_{0}x$, forward pass yields:
$$
h = W_{0}x + \Delta W x = W_{0}x + W_{B} W_{A} x
$$
Based on loRA, QLoRA \cite{dettmers2023qlora} introduced 4-bit NormalFloat, Double Quantization and Paged Optimizers, which further reduce memory requirements and improve training efficiency. Inspired by these fine-tuned approaches, we follow methods like QLoRA \cite{dettmers2023qlora} as a low-cost way to inject knowledge into LLMs. 

\paragraph{Prompt Engineering}
The prompt is a crucial part of fine-tuning. People usually maximize the potential of LLMs for downstream tasks by designing prompts. Notable works like CoT \cite{wei2022chain}, augment each exemplar in few-shot prompting with a chain of thought for an associated answer. zero-shot \cite{wei2022semantic} and few-shot learning \cite{wang2020generalizing}  bring improvements to LLMs in the downstream task, which could lead to state-of-the-art effects without any training. For medical questions and answering tasks, designing a suitable prompt is critical to success. 

\section{Methods and Experiments}

\subsection{DataSets for Medical Knowledge}
The data for knowledge injection consists of medical text collected from various resources on the internet, including text from real conversations and generated conversations from ChatGPT. It comprises a total of 116.41K samples, with 15.61M tokens in total. These conversations cover a wide range of medical topics, including disease diagnosis, treatment advice, medication consultation, and more \ref{table-datasets}. 

\begin{table*}[b]
    \centering
    \caption{Composition of datasets and sources.}
    \label{table-datasets}
    \begin{tabular}{lcccrrr}
    \toprule
        \textbf{Type} & \textbf{Source} & \textbf{Number of samples} \\ \midrule
        Real conversations & HealthCareMagic.com & 100K \\ 
        Real conversations  & icliniq.com & 10K \\
        Generated conversations & ChatGPT & 5K \\
        Real conversations & BI55/MedText(Huggingface) & 1.41K \\
    \bottomrule
    \end{tabular}
\end{table*}

\subsection{Down-stream Instruction Tuning}
\begin{figure}[htbp]
\centering
\includegraphics[scale=0.4]{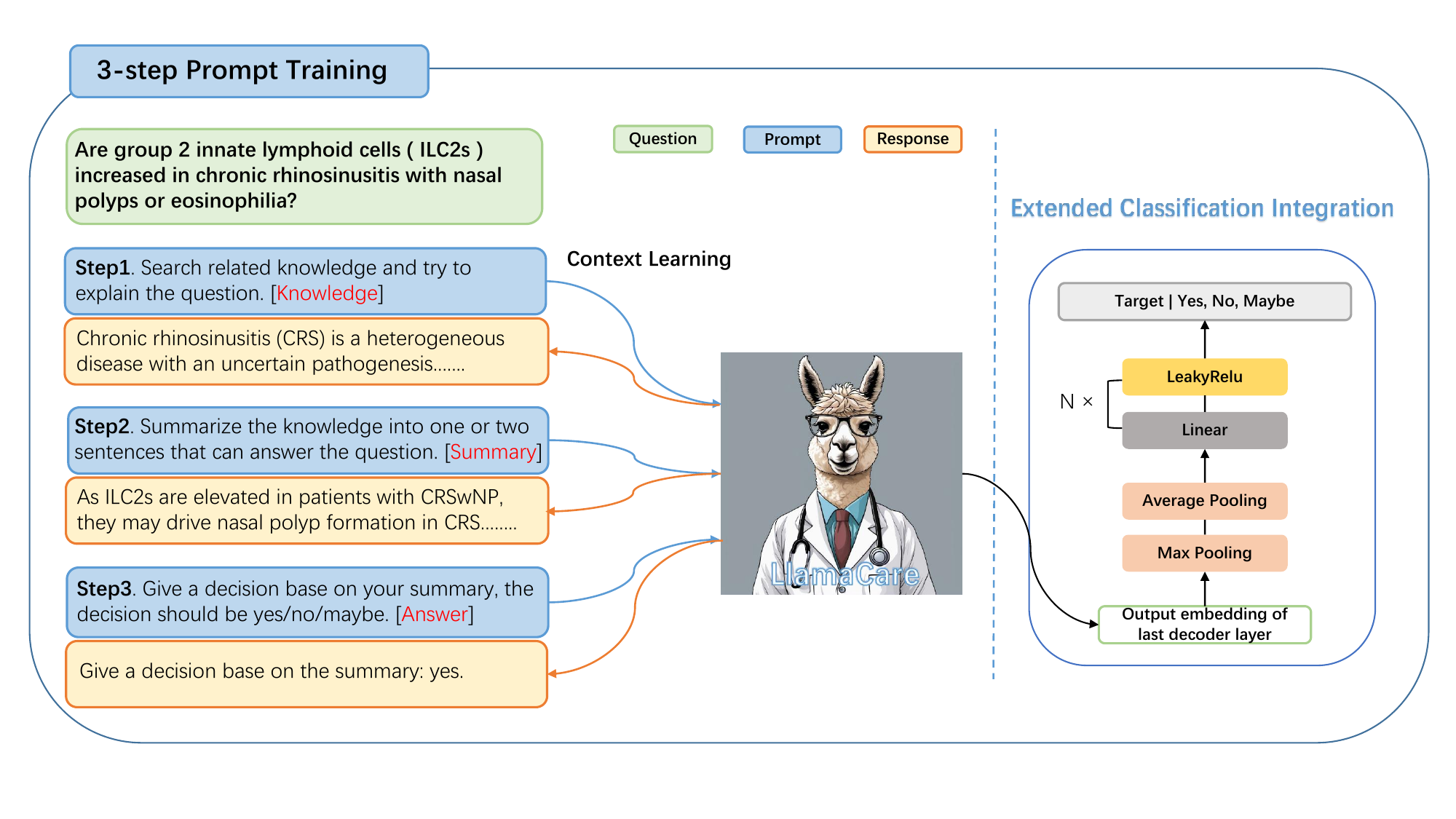}
\caption{\textbf{Left part.} This prompt motivates the model to think about the knowledge related to the question as well as extrapolating answers from knowledge. Fine-tuning LLMs by this prompt can avoid the phenomenon of the model reciting the answer. \textbf{Right part.} The ECI module allows the model to output an additional label each time during the training.}

\label{figure2}
\end{figure}

In addition to enhancing the conversational capabilities of the model, we aim to enhance our model by incorporating professional medical knowledge and reasoning abilities. We utilize training sets from open-source medical choice question-answering datasets PubMedQA \cite{jin2019pubmedqa}. And we want models to learn problem-solving techniques in this exam. We designed a 3-step prompt for learning problem-solving techniques, as shown in Figure  \ref{figure2}. This prompt allows the model to learn the answer and associated knowledge at the same time during the training process, instead of just reciting the answers.

\subsection{Extended Classification Integration}
It is a trigger problem that sometimes LLMs can not follow our instructions to do classification problems simply as we desire. Even if it has been instruction-tuned first. For instance, we want LlaMA-2-13b-chat response to a classification label only, we use a zero-shot prompt like “Please give an option in the choice and don’t explain it.” However, sometimes the model adds explanations with the label and sometimes it refuses to answer the question. If we evaluate classification tasks on a benchmark, we need to manually check whether some answers are correct. To handle this problem, we propose Extend Classification Integration for LLMs. Extend Classification Integration is a classification network design to give an additional classification result. Different from classifying each position of embedding to the vocabulary in the head of the decoder, ECI classifies the whole embedding to a label we want as right in Figure \ref{figure2}.

Specifically, we utilize the output embedding of the last decoder layer in LLaMA-2 as the input for ECI. We have designed the network structure of ECI. Due to the excessive dimensionality of the output embedding, the introduction of excessive parameters by ECI is observed. To address the problem, we initially used pooling to lower the dimensionality. In LLaMA-2-13b, the dimension of the output is [b,s,5120], where b is the batch size, and s is the max sequence length. s=1900 in our experiments, after flattening the number of neurons, the input layer would be $5120 \times 1900 = 9728000$. Thus if we use linear for dimension reduction directly, the parameters here would be $9728000 \times outputdimension$. The first layer in the neural network could bring billions or even more than 10 billion of parametric quantities. Thus, we use max pooling and average pooling to lower the dimension and extract features first. Then, there are several linear layers and activate functions. In our experiments, N = 5 and K = 8 for the kernel of max pooling and average Pooling.

Furthermore, we introduced the loss of ECI into the fine-tuning process, allowing for simultaneous updating of ECI and the adapter simultaneous. The loss function for ECI is cross-entropy loss and we use  $\lambda$ to adjust the weight of loss between text generation and ECI. Formally, we define our loss as follows: 
$$
L = (1 - \lambda) l_{TextGeneration} +  \lambda l_{ECI},
$$
$$
L = (1 - \lambda)\sum_{i=1}^{N}l_{CrossEntropy}(y(0)^{true}_{i}, y(0)^{pred}_{i}) + \lambda l_{CrossEntropy}(y(1)^{true}, y(1)^{pred})
$$
Where $y(0)$ means the output of the text-generation head and $y(1)$ means the output of ECI. The loss of text generation treats each position of the embedding vector as a separate classification task and $i$ is the position of the embedding sequence. $y^{true}$ and $y^{pred}$ is the label and prediction of corresponding index. Our loss considers text generation and classification simultaneous.

We use AdamW \cite{loshchilov2017decoupled} as the optimizer and linear learning rate with a start from $5e-5$. R=16 for LoRA rank and freeze all other parameters except the Qproj layer, Vproj layer and ECI module. Other hyper-parameters can be seen in our GitHub repository.

\section{Evaluation} 

\paragraph{BLEU Score}
The BLEU score \cite{papineni2002bleu} was first used in machine translation tasks and now is a widely adopted metric in text generation, and question and answering evaluation. It quantifies the quality of generated text by comparing them to one or more references in ground truth. BLEU calculates the similarity between the generated text and the reference text based on n-grams and their respective counts. A higher BLEU score signifies a higher degree of alignment with the reference translations, indicating better translation quality. The calculation of the BLEU score is defined as follows. We use BLEU-4 in our evaluation.
$$
\begin{aligned}
& BLEU-N= \frac{\sum_{C \in\{\text { Candidates }\}} \sum_{\text {n-gram } \in \mathcal{C}} \operatorname{Count}_{\text {clip }}(\text { n-gram })}{\sum_{C^{\prime} \in\{\text { Candidates }\}} \sum_{\text {n-gram }{ }^{\prime} \in C^{\prime}} \operatorname{Count}\left(\text { n-gram }{ }^{\prime}\right)} \\
\end{aligned}
$$

\paragraph{ROUGE Score}
The ROUGE score \cite{chin2004rouge} is a set of evaluation metrics commonly employed in summarization and text generation tasks. It measures the overlap between the generated text and reference summaries using diverse algorithms. ROUGE scores encompass various measures such as ROUGE-N (evaluating n-gram overlap), ROUGE-L (assessing the longest common subsequence), and ROUGE-S (measuring skip-bigram overlap), among others. ROUGE scores provide insights into the quality and adequacy of generated summaries. The ROUGE score is defined as follows. We use ROUGE-1 in our evaluation.

$$
Rouge\small{-N} = \frac {\sum \limits_{S \in ReferenceSummaries}\sum \limits_{gram_n \in S} Count_{match}(gram_n)} {\sum \limits_{S \in ReferenceSummaries}\sum \limits_{gram_n \in S} Count(gram_n)}
$$

\paragraph{Human Evaluation}
In order to evaluate the model's responses naturally, we introduced human evaluation. We invited 3 doctors to score the generated samples and to eliminate subjectivity, we gave a rating from 1-10 and finally took the average. Human evaluation captures subjective nuances that automated metrics may overlook, providing valuable insights into the text's quality from a human perspective.

\paragraph{ChatGPT Evaluation}
ChatGPT is widely used to evaluate other LLMs. Using ChatGPT for evaluation provides a valuable way to incorporate human preferences without direct human involvement. As ChatGPT is trained with RLHF, it learns human preferences to a significant extent. This makes it a reliable tool for assessing and rating other large models. Leveraging ChatGPT's conversational capabilities allows for an effective evaluation of models while considering human-like interactions and preferences.

\section{Baseline}

\paragraph{LLaMA-2 \cite{touvron2023llama2} }
is an enhanced iteration of LLaMA that has further fine-tuning with instructions. LLaMA-2 is a collection of pre-trained and fine-tuned large language models ranging in scale from 7 billion to 70 billion parameters. Llama2-Chat is an instruction-tuned model, which optimized for dialogue use cases.

\paragraph{ChatGPT}
is a commercially released language model by OpenAI, introduced in November 2022. It has exhibited exceptional performance across a broad spectrum of natural language processing tasks, spanning various domains, including the field of medicine. While specific details regarding ChatGPT are proprietary, it is generally presumed that the size of ChatGPT is roughly 175B, the same as GPT-3. \cite{wu2023pmc}

\paragraph{Med-Alpaca \cite{han2023medalpaca}}
is a language model that underwent additional fine-tuning on Alpaca \cite{taori2023stanford} using medical instruction data. Med-Alpaca aims to enhance its performance in medical dialogues and question-answering tasks. 

\paragraph{Chat-Doctor \cite{li2023chatdoctor}}
is a fine-tuned medical language model based on LLaMA. It was trained on Wikipedia and the disease database. Chat-Doctor has shown great performance on many benchmarks.

\section{Benchmarks}

\paragraph{PubMedQA \cite{jin2019pubmedqa}}
 is a biomedical question-answering dataset. The primary objective of PubMedQA is to provide answers to questions in the choice of yes/no/maybe. The dataset consists of three subsets: 1k manually labeled pairs (PQA-L), 61.2k unlabeled pairs (PQA-U), and 211.3k artificially generated pairs (PQA-A). We take PQA-A as the training set, PQA-L serves as the test set, and the PQA-U portion is disregarded.

\paragraph{USMLE-step-1-3}
USMLE (United States Medical Licensing Examination) is a professional qualification examination for medical licensing. The examination aims to assess the physician's ability to apply this knowledge and skills comprehensively, ensuring the provision of optimal healthcare outcomes. It includes 3 steps. We collected the official example papers from step1, step2, and step3 to format one of our evaluation benchmarks. We removed the question with the images because the LLMs cannot extract information from pictures. The detail of the USMLE benchmark can be seen in Table \ref{tab-usmle}.

\begin{table*}[th]
    \centering
    \caption{USMLE Sample Question Step 1-3}
    \label{tab-usmle}
    \begin{tabular}{lcccrrr}
    \toprule
        \textbf{Name of Exam Paper} & \textbf{Number of Questions} \\ \midrule
        Sample test question-step1 & 94 \\ 
        Sample test question-step2  & 109 \\
        Sample test question-step3 & 122 \\
        Total & 325 \\
    \bottomrule
    \end{tabular}
\end{table*}

\section{Results}
We executed the above experiments to analyze and compare LlamaCare with baseline models on multiple dimensions. The results of the experiment can be seen in Table \ref{tab-Ablation study}  and Table \ref{tab-benchmark}.

\begin{table}[th]
    \centering
    \caption{Ablation study on BLEU-4, ROUGE-1 and PubMedQA, evaluated by accuracy score. "3SP" means "3-step Prompt". Note that in LlamaCare-one-shot-3sp-ECI, the ECI module brings about 100M additional parameters.}
    \label{tab-Ablation study}
    \resizebox{\textwidth}{!}{
        \begin{tabular}{lcccrrr}
            \toprule
                \textbf{Methods} & \textbf{Size} & \textbf{Quantization} & \textbf{Fine-tuning} & \textbf{BLEU-4} & \textbf{ROUGE-1} & \textbf{PubMedQA (benchmark)} \\ 
                \midrule
                LLaMA-2 (Baseline) & 13B & 4 bit & \XSolidBrush & \XSolidBrush & \XSolidBrush & 52.40 \\ 
                LLaMA-2-3SP & 13B & 4 bit & \XSolidBrush & 43.22 & 34.43 & 46.80 \\
                LLaMA-2-one-shot-3SP & 13B & 4 bit & \XSolidBrush & 46.48 & 35.53 & 53.40 \\
                LLaMA-2-few-shot-3SP & 13B & 4 bit & \XSolidBrush & 42.97 & 30.58 & 54.20 \\
                LlamaCare-one-shot-3SP & 13B & 4 bit & \CheckmarkBold & \textbf{47.45} & \textbf{46.42} & \textbf{57.60} \\
                LlamaCare-one-shot-3SP-ECI & 13.1B & 4 bit & \CheckmarkBold & \XSolidBrush & \XSolidBrush & \textbf{55.20} \\
            \bottomrule
        \end{tabular}
    }
\end{table}

\begin{table}[th]
    \centering
    \caption{Evaluation on the benchmark. Quantization results in a reduction in the model's computational precision, where 4-bit quantization means the model's computational precision decreases from 32-bit floating numbers to 4-bit floating numbers.}
    \resizebox{\textwidth}{!}{
    \label{tab-benchmark}
    \begin{tabular}{lcccrrr}
    \toprule
        \textbf{Methods} & \textbf{Model size} & \textbf{Quantization} & \textbf{Fine-tuning} & \textbf{USMLE} & \textbf{PubMedQA} & \textbf{Human Evaluation} \\ \midrule
        LLaMA-2 (Baseline) & 13B & 4 bit & \XSolidBrush & 32.30 & 52.40 & 7.3 \\ 
        ChatGPT & 175B &  \XSolidBrush & \CheckmarkBold & \textbf{58.76} & \textbf{63.90} & \textbf{8.8} \\
        Med-Alpaca & 13B & \XSolidBrush & \XSolidBrush & 45.63 & 53.20 & 6.8 \\
        Chat-Doctor & 7B & \XSolidBrush & \CheckmarkBold & 41.87 & 54.30 & 7.6 \\
        LlamaCare & 13B & 4 bit & \CheckmarkBold & 44.64 & \textbf{57.60} & \textbf{8.5} \\
    \bottomrule
    \end{tabular}
    }
\end{table}

From the ablation study, our prompt brings some improvements in one-shot learning and few-shot learning, where n=3 in few-shot. The prompt activated LLaMA-2’s ability of question-understanding and question-solving on PubMedQA which led to an increase from 52.4\% to 53.4\% and 54.2\% respectively. Besides, evaluation on BLEU-4 and ROUGE-1 both increased greatly by the prompt, which shows that the model is thinking the same way as we want. The prompt is also valid in fine-tuning since it boosts the accuracy to 57.6\%. However, it is a fly in the ointment that ECI has not been able to improve the classification accuracy compared to text generation in the case of fine-tuning as we desire. We speculate this is due to the high input dimension of the classification network, which is so wide that it would be relatively difficult to extract effective features during training. The question can be progressively explored, i.e. different backbones. In a sense, ECI exceeded zero-shot and one-shot learning and solved the problem of models not answering classification results by instruction. It deserves to be adopted in future studies.

In benchmark evaluation, LlamaCare stands out as the most remarkable performance in PubMedQA and human evaluation among models with similar parameter sizes, even with the adoption of 4-bit quantization. In comparison to the baseline models, we achieved better performance while consuming less carbon emissions. We also gave some responses to LlamaCare in diagnostic questions, which shows LlamaCare could capture the key point of the symptoms, please see details in the Appendix.

\section{Conclusion}
In this paper, we first analyzed the challenges and limitations faced by current state-of-the-art LLMs in the medical domain. Subsequently, we fine-tuned LlamaCare using a low-carbon approach to medical question-answering datasets and pathology datasets. We introduced a three-step prompt that significantly improved accuracy in both one-shot and few-shot scenarios. Furthermore, we proposed an Extended Classification Integration (ECI) to address the issue of LLMs not following instructions perfectly when doing classification problems after instruction tuning, which holds potential for future research. LlamaCare achieved an advanced performance level compared to models with similar parameter sizes on benchmarks.





\bibliographystyle{plain}
\bibliography{llamacare}

\newpage

\section{Appendix}

\begin{figure}[htbp]
\centering
\includegraphics[width=\textwidth]{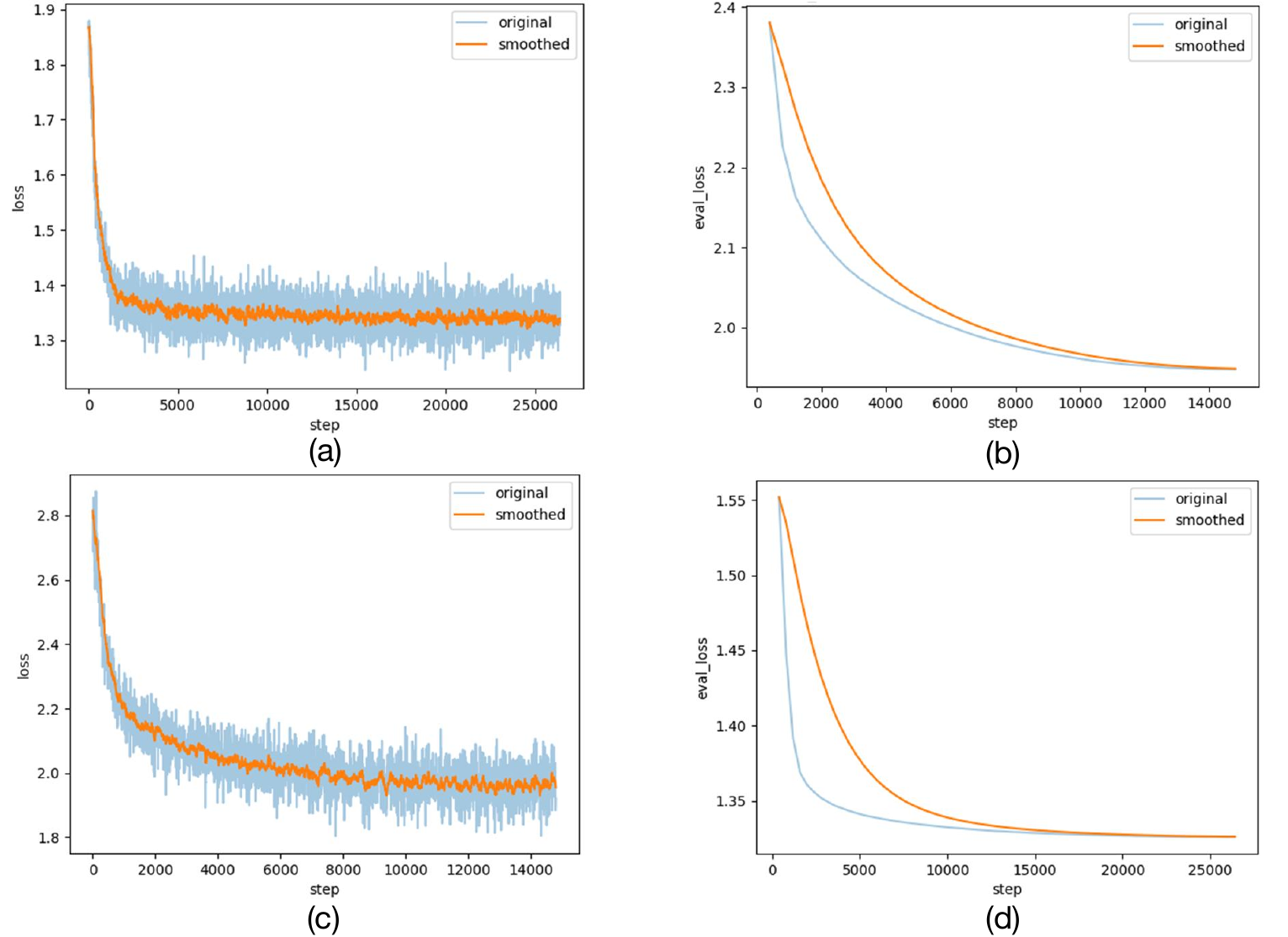}
\caption{Loss of fine-tuning in experiments. (a) Fine-tune on medical text. (b) Evaluation of medical text. (c) Fine-tuning on benchmark datasets. (d) Evaluation of benchmark datasets. }
\end{figure}

\subsection{LlamaCare's inference effect on real medical consultation}

\paragraph{Question} A 50-year-old male presents with a history of recurrent kidney stones and osteopenia. He has been taking high-dose vitamin D supplements due to a previous diagnosis of vitamin D deficiency. Laboratory results reveal hypercalcemia and hypercalciuria. What is the likely diagnosis, and what is the treatment?
\paragraph{Response} The likely diagnosis is vitamin D-dependent hypercalcemia. Vitamin D is a fat-soluble vitamin that is converted to its active form, calcitriol, in the kidneys. Calcitriol is involved in the regulation of calcium metabolism and is responsible for maintaining normal levels of calcium in the blood. When vitamin D is taken in excess, it can lead to an overproduction of calcitriol, which can cause hypercalcemia and hypercalciuria. Treatment includes stopping vitamin D supplements and initiating treatment with a diuretic to help remove excess calcium from the body. Additionally, a low-calcium diet may be recommended to help reduce the amount of calcium in the blood. In severe cases, medications such as bisphosphonates may be prescribed to help lower calcium levels.

\paragraph{Question} A 7-year-old boy presents with a fever, headache, and severe earache. He also complains of dizziness and a spinning sensation. Examination reveals a red, bulging tympanic membrane. What are the differential diagnoses, and what should be done next?
\paragraph{Response}
Based on the patient's symptoms and examination findings, the differential diagnoses for the patient's condition include: Acute otitis media (AOM): AOM is an infection of the middle ear that can cause fever, headache, earache, dizziness, and a spinning sensation. Meningitis: Meningitis is an infection of the meninges that can cause fever, headache, dizziness, and a spinning sensation.Labyrinthitis: Labyrinthitis is an inner ear infection that can cause dizziness, vertigo, and hearing loss.Viral gastroenteritis: Viral gastroenteritis can cause fever, headache, and dizziness. The next step in management would be to obtain a tympanocentesis to culture the middle ear fluid and determine if there is a bacterial infection present. If the culture is positive, antibiotics will be prescribed. If the culture is negative, further evaluation for meningitis or viral gastroenteritis would be needed.

\paragraph{Question}A 35-year-old woman presents with a persistent dry cough, shortness of breath, and fatigue. She is initially suspected of having asthma, but her spirometry results do not improve with bronchodilators. What could be the diagnosis?
\paragraph{Response}
Based on the patient's symptoms and negative response to bronchodilators, the diagnosis could be chronic obstructive pulmonary disease (COPD). COPD is a chronic inflammatory condition of the airways that can cause symptoms similar to asthma, including cough, shortness of breath, and fatigue. However, unlike asthma, COPD does not respond to bronchodilators and may require long-term treatment with inhaled corticosteroids and bronchodilators. Other possible diagnoses include chronic bronchitis, emphysema, or pulmonary fibrosis. A complete diagnostic workup, including chest X-ray, pulmonary function tests, and CT scan, would be needed to confirm the diagnosis and determine the appropriate treatment.

\paragraph{Question} A 50-year-old male presents with severe abdominal pain, vomiting, and constipation. He has a history of long-standing hernia. On examination, the hernia is tender, firm, and non-reducible. What's the likely diagnosis and the next steps?
\paragraph{Response}
The likely diagnosis is strangulated hernia. A strangulated hernia occurs when the hernia becomes trapped and the blood supply to the intestine is cut off, leading to ischemia and necrosis of the intestinal tissue. This is a surgical emergency and requires immediate attention. The next steps would be:
Emergency surgical exploration to reduce the hernia and relieve the strangulation.Repair of the hernia using a mesh or primary repair.
Management of any associated complications such as bowel ischemia, perforation, or abscess.Prolonged post-operative recovery and close monitoring for any signs of complications.

\paragraph{Question}
A newborn baby presents with eye redness and a thick purulent discharge in both eyes. The mother has a history of untreated chlamydia. What could be the cause?
\paragraph{Response}
Based on the symptoms and history of the baby, the most likely cause is a congenital infection such as chlamydia. Congenital infections are infections that are acquired by the baby during pregnancy or childbirth. They can be caused by bacteria, viruses, or other microorganisms. Chlamydia is a common cause of congenital infections. It is important to have the baby tested for chlamydia and treated promptly. The baby may need antibiotic eye drops or oral antibiotics. The mother should also be tested for chlamydia and treated if she is positive.

\end{document}